# Online Stroke and Akshara Recognition GUI in Assamese Language Using Hidden Markov Model

**SRM Prasanna, Rituparna Devi, Deepjoy Das, Subhankar Ghosh, Krishna Naik** *

* Department of EEE, Indian Institute of Technology Guwahati

***Abstract***-The work describes the development of Online Assamese Stroke & Akshara Recognizer based on a set of language rules. In handwriting literature strokes are composed of two coordinate trace in between pen down and pen up labels. The Assamese aksharas are combination of a number of strokes, the maximum number of strokes taken to make a combination being eight. Based on these combinations eight language rule models have been made which are used to test if a set of strokes form a valid akshara. A Hidden Markov Model is used to train 181 different stroke patterns which generates a model used during stroke level testing. Akshara level testing is performed by integrating a GUI (provided by CDAC-Pune) with the Binaries of HTK toolkit classifier, HMM train model and the language rules using a dynamic linked library (dll). We have got a stroke level performance of 94.14% and akshara level performance of 84.2%.

## I. INTRODUCTION

Online Handwriting recognition also termed as character recognition is the task of recognizing written message by processing handwritten data. For development of automatic handwriting recognition system a basic unit has to be selected. In Assamese handwriting recognition the components of the Assamese symbols are taken as the basic units and are commonly termed as strokes. For online handwriting recognition the two coordinates trace composed between one pen down to one pen up is termed as stroke. For Assamese handwriting recognition a set of 181 strokes were identified as valid strokes. An akshara may comprise of a single stroke or multiple strokes from the set of 181 strokes. Analyzing the data from a large number of writers it was found that almost all Assamese aksharas can be written as a combination of maximum 8 strokes based on which 8 language rule models have been created using which a symbol written by an user can be identified.

### 1.1 Assamese Language

Assamese is the easternmost Indo-Aryan language mainly used in the state of Assam in North-East India and is the official language of this state, spoken by around 13 million people. Recognition of handwritten Assamese script is complicated as it is primarily a cursive language with a large number of ligatures formed by the combination of potentially all the consonants with one another. Vowels can either be independent or dependent upon a consonant or a consonant cluster. The phonetic character set and behavior of Assamese is derived from Sanskrit and it is written using the Assamese script. The script presently has a total of 11 vowel letters, around 41 consonants, 10 numerals and a number of conjuncts, vowel modifiers, consonant modifiers and other symbols. The vowels, consonants and numerals are already defined in the language. However frequency of using conjuncts is not well defined and hence the Assamese script has been carefully analyzed to prepare a list of valid conjuncts.

### 1.2 Assamese Handwritten symbol database

The preparation of Assamese database begins with the scanning of the words in the pages of Hemkosh Dictionary by Hem Chandra Barua which yields 185 conjuncts. Meanwhile, a list of 147 conjuncts prepared by the Resource Centre for Indian Language Technology Solutions (RCILTS), IITG is decoded after discussing it with Mr. Pallav Barua, RCILTS, and IIT Guwahati. This document when compared with our document on list of conjuncts gives us another list of commonly used conjuncts. Thus, the Assamese script of 241 aksharas is formed consisting of 11 vowels, 41 consonants, 147 conjuncts, 10 numerals, 10 vowel modifiers, 2 consonant modifiers and 20 additional symbols, which are then sent for data collection. The list of 241 symbols is depicted in Figure 1.

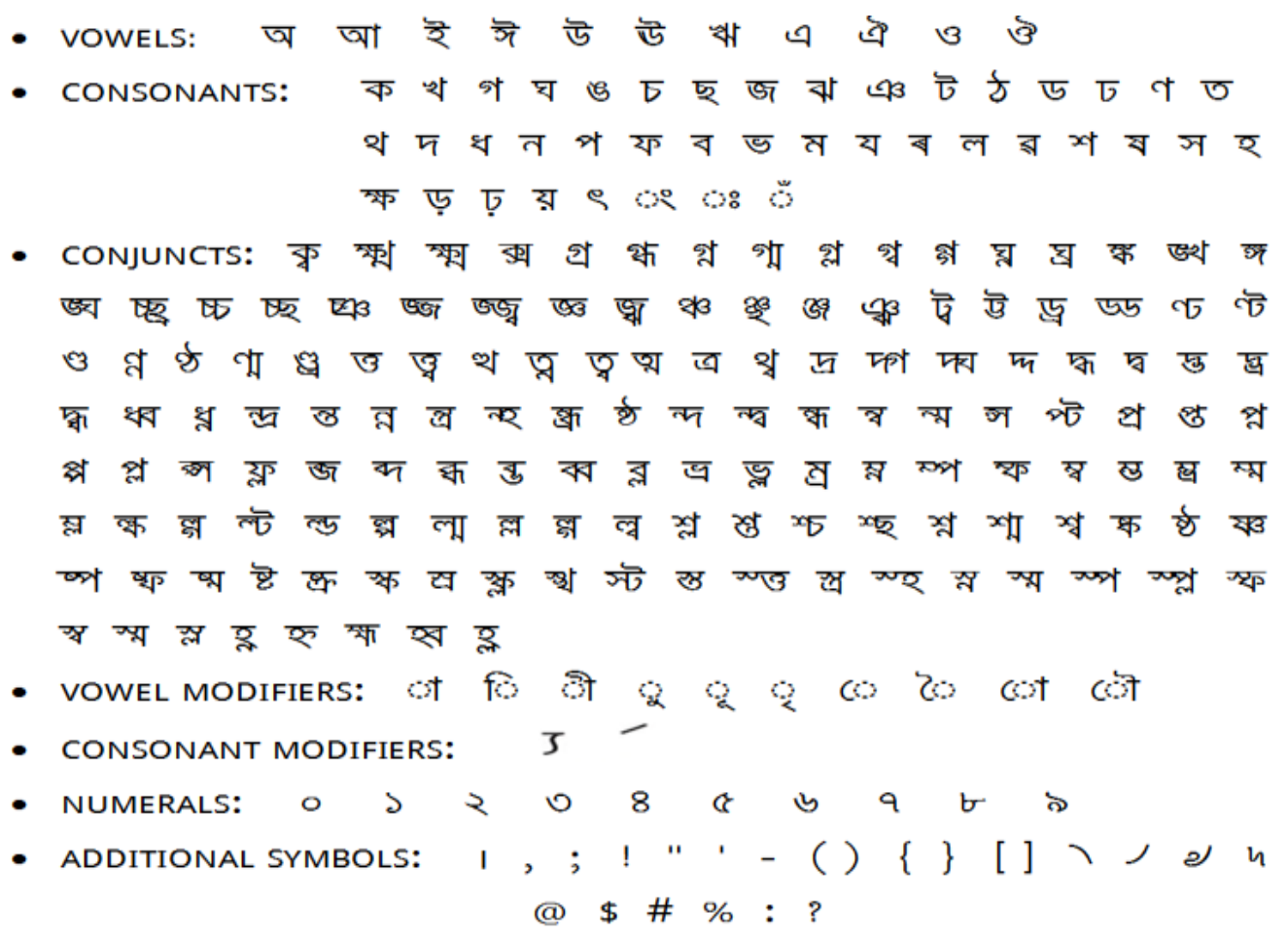

**Figure 1: 241 Assamese Aksharas**

Tablet PC is used to collect data with a sampling rate of 120 Hz. The captured information contains the sampled x, y coordinate values along with the writer information. Based on writer feedback a final list of 147 Assamese symbols is created which consists of 11 vowels, 41 consonants, 55 conjuncts, 10 numerals, 23 special symbols and 5 characters whose shape change on addition of modifiers. The list of 147 symbols is shown in Figure 2.

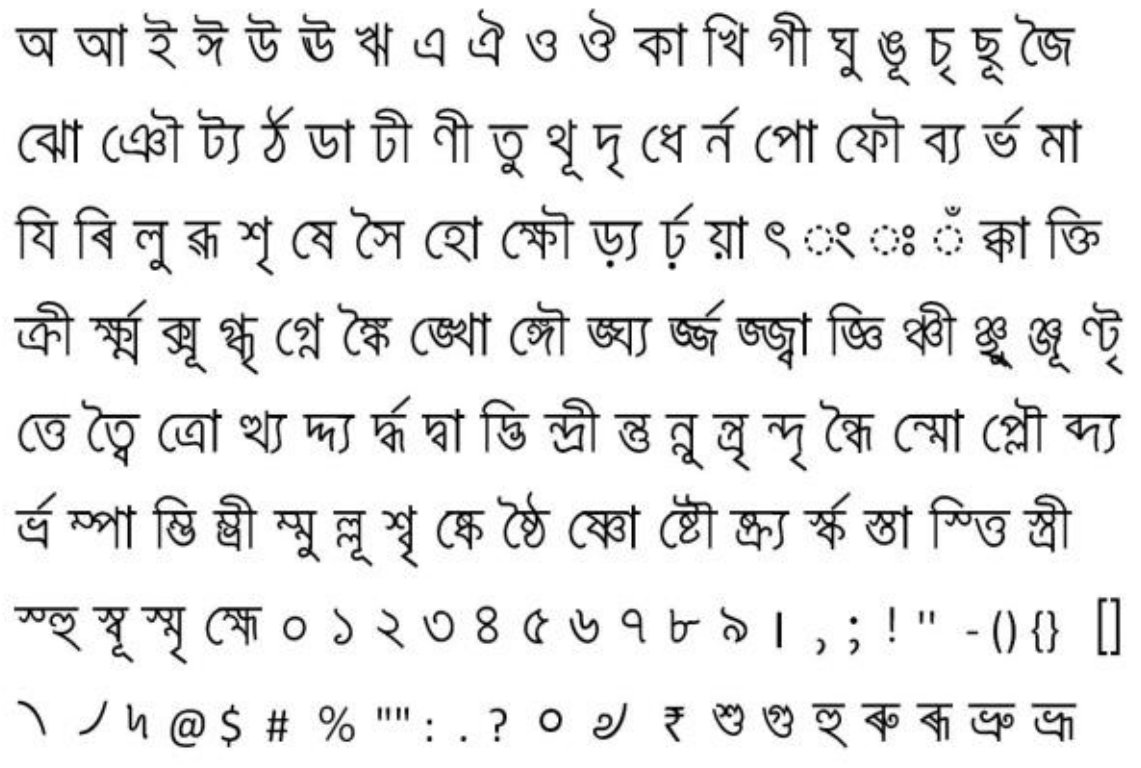

**Figure 2: 147 isolated aksharas**

### 1.3 Basic components of Assamese handwriting

The basic component of aksharas is termed as strokes. In handwritten literature the coordinate trace between one pen down to pen up is termed as stroke. However in case of Assamese handwriting which is a case of cursive handwriting several components may be merged in one pen down to its pen up. Some components may also be broken by infrequent writers. So data grouping has been done in 3 different ways namely strokes, substrokes and suprastrokes. The Class-A data or strokes consists of a typical basic component of aksharas agreed naturally by majority of non-cursive writers. The Class-B data or substrokes contained components are formed by merging several components or strokes. There are 263 distinct substrokes. Again some infrequent writers break the components of the .stroke into more than one component. The splitted components form the Class-C data or suprastrokes. There are 104 distinct suprastrokes. A list is prepared combining all the above strokes of different data groups and we have a final list of 203 distinct strokes in Assamese handwriting. The final strokes list is depicted in Figure 3.

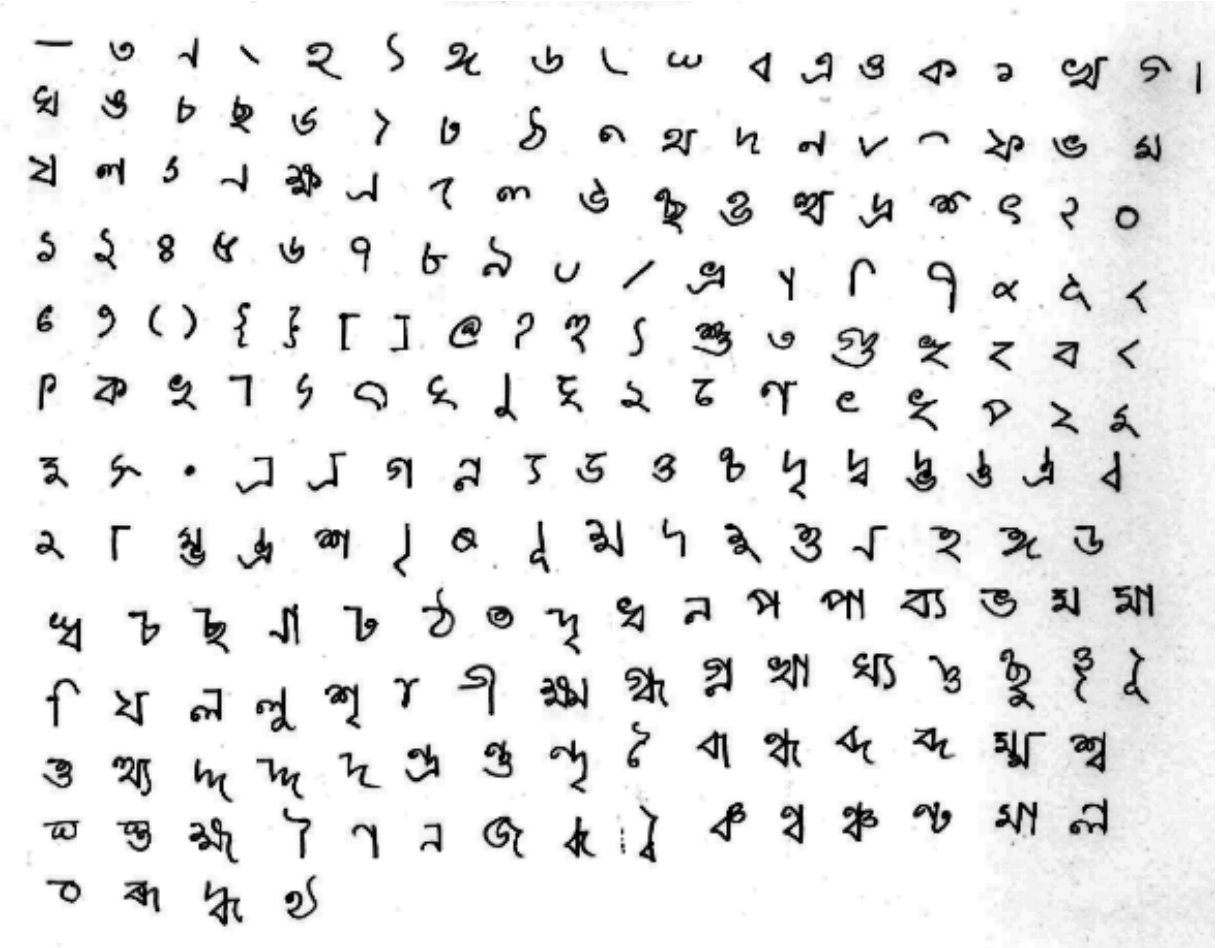

**Figure 3: 203 isolated assamese strokes**

## II. AKSHARA RECOGNITION SYSTEM

The schematic block diagram of Online Assamese Handwriting Recognition system is shown in Figure 4.

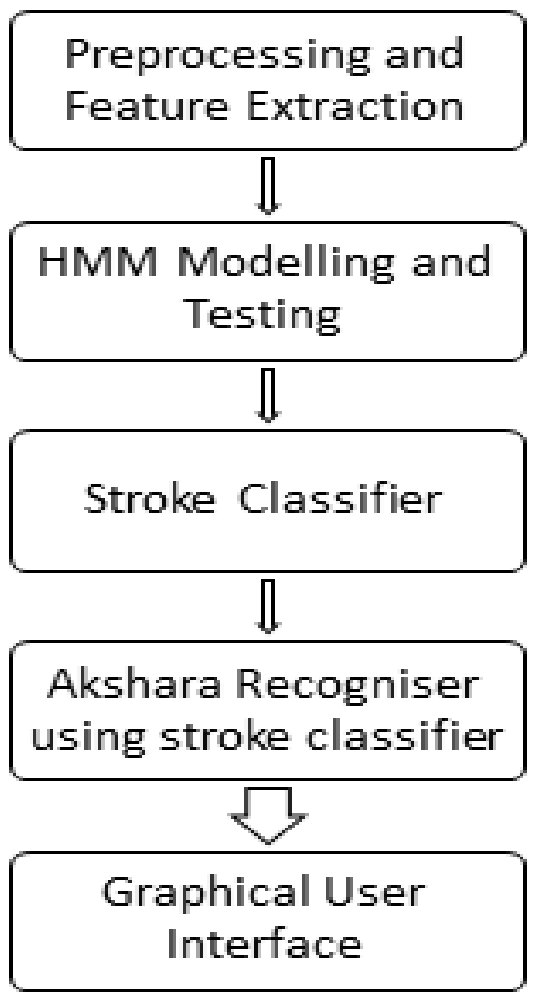

**Figure 4: Block Diagram of Recognition System**

### 2.1 Preprocessing

In online mode, the handwritten pattern is captured as a series of (x, y) coordinates. The preprocessing stage performs size normalization, smoothing, interpolation of missing points, removal of duplicate points, and resampling of the captured coordinates [1].

#### 2.1.1 Size Normalisation

Size normalization normalizes the variations in the size of the pattern, due to the various writing styles and size of the box provided for writing. It is performed by scaling each pattern both horizontally and vertically [2]

#### 2.1.2 Smoothing

Smoothing eliminates the noise captured during the data collection process. It is performed using a moving average filter of size three. Each pattern is smoothed both in x and y directions separately

#### 2.1.3 Removal of duplicate points

Duplicate points results in data redundancy and does not contain any information for recognition. These points are removed before feature extraction

#### 2.1.4 Resampling

Resampling removes the variations in the data due to the writing speed of the writers. It is performed by linear interpolation of missing points which results in a sequence of equidistant points

### 2.2 Feature Extraction

The preprocessed x, y coordinates, first and second derivatives of x, y coordinates are considered as features. The first and second derivatives interprets the change and change of change in x, y coordinates, respectively. The significance of calculating first derivative is to observe the change in the trajectory at current point. A window size of two is considered for calculating first derivative. The second derivative is calculated in order to examine the change of change in trajectory at current point. Here also, a window size of two is considered.

### 2.3 HMM Modelling and Testing

The variations in each stroke class are modeled using Hidden Markov models (HMM). One HMM is constructed for each stroke, by training on example strokes. Seven state HMM is used to test and train the models which is determined experimentally. The number of Gaussian mixture is optimized to the better recognition accuracy. HTK tool is used to test and train the models [3].

HMM models a doubly stochastic process, one observable and the other hidden [4]. The observable stochastic process contains information about the hidden stochastic process. In our work, the sequence of feature vectors from the online handwriting is the observable stochastic process and the underlying hand movement is the hidden stochastic process. The basic assumption for using HMM for handwriting recognition is that there are unique handwriting movements for writing each of the basic components, namely stroke [5]. Further, the left to right structure is used assuming unique directions of handwriting movements. In the present work, one left to right, continuous density HMM is developed for modelling each stroke.

#### 2.3.1 Training

A HMM consists of a set of states and the transitions associated with it and are trained using Baum-Welch re-estimation or expectation maximization (EM) approach [4]. The procedure starts with some initial model and improved model parameters are re-estimated using the given set of feature vectors. For the next iteration, the most recent model is the initial model and again re-estimation is done using the same set of feature vectors. This process is repeated until model parameters become static and the model of the last iteration is stored as the model for the given class.

#### 2.3.2 Testing

Testing involves finding out the class information for the examples that are unknown to the trained model. The likelihood probability of the given testing example against each of the trained HMM models are found out and the model with the highest likelihood is hypothesized as the class. The process is repeated for all the testing examples and the class information is noted.

### 2.4 Stroke Classifier

The basic block diagram of the stroke recognizer is shown in Figure 5.

The stroke classifier has been built using HMM modelling technique. 203 HMM models have been built for each of the 203 strokes finalized during script analysis prior to which the database is annotated with the annotation tool at stroke level and six dimensional features are extracted from the preprocessed coordinates .All the test examples corresponding to each stroke class are tested against all the stroke models. If misclassification arises due to similarity in shape between two strokes those stroke classes are merged stroke classifier performance is evaluated. A stroke classifier with 181 HMM models is finally developed.

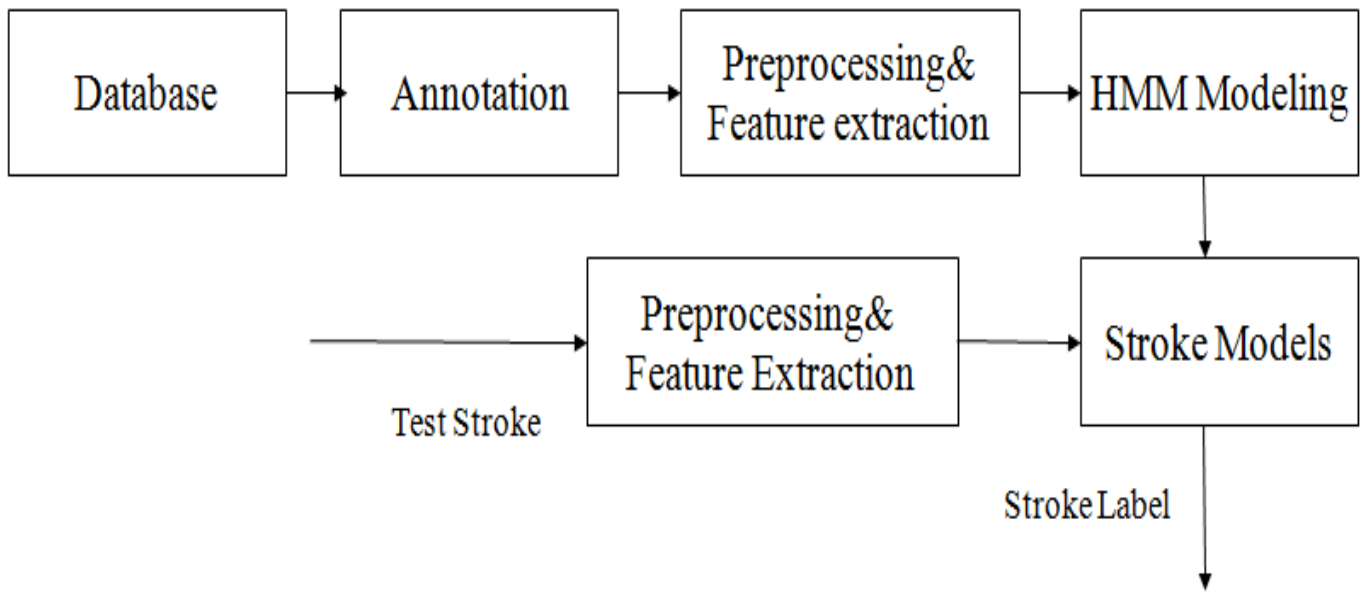

**Figure 5: Block diagram of stroke classifier**

### 2.5 Akshara Recogniser using HMM Stroke Classifier

The basic block diagram of akshara recognizer is shown in Figure 6.

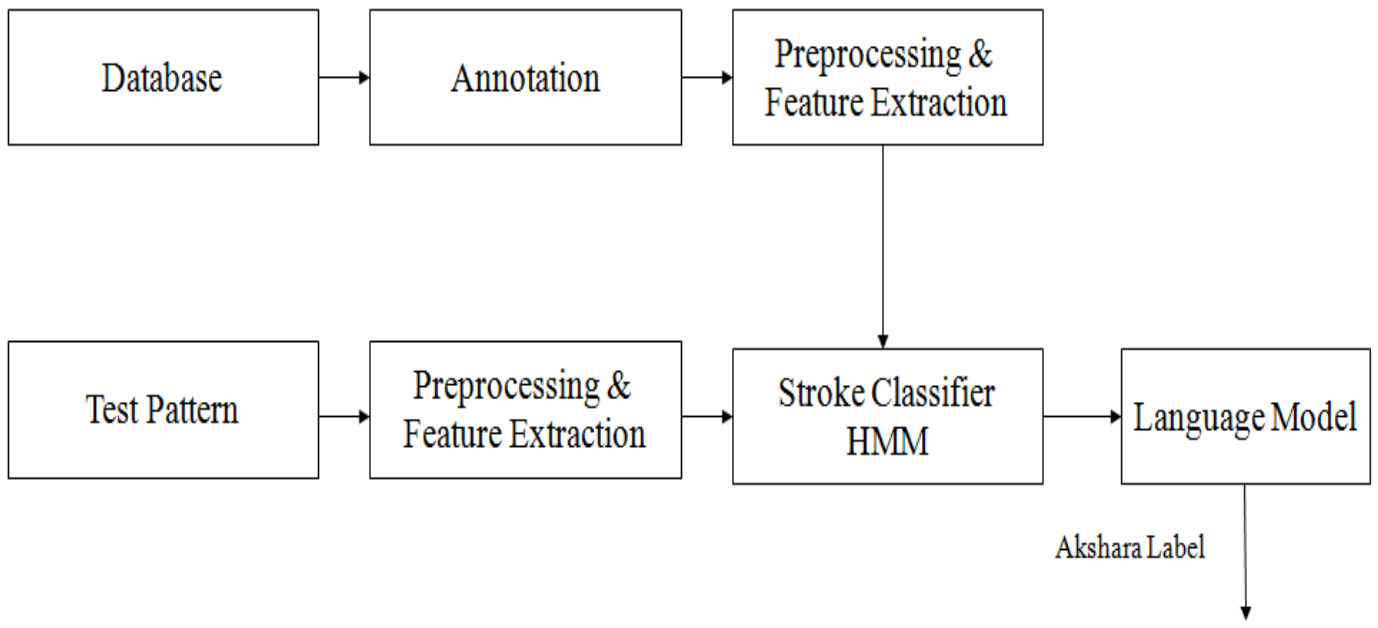

**Figure 6: Block diagram of Akshara Recognizer**

An Akshara is a combination of different strokes. During testing of akshara, the strokes of an akshara are preprocessed and the features are extracted. The features of preprocessed coordinates are then tested against the developed stroke classifier. The recognized stroke labels along with the language models are used for the recognition of an akshara. Language model refers to the combination of strokes which form the akshara. The isolated Akshara recognizer makes use of a language rule model for the recognition of the akshara label. First the strokes of the akshara are recognized and then the akshara, using the language rule. If the stroke classifier recognizes all the strokes correctly then only we consider it as recognized akshara.

#### 2.5.1 Language Rule for Akshara Recogniser

An akshara can be a single stroke or a combination of multiple strokes. Also the same akshara can be written as a combination of two strokes, three strokes, four stroke and so on.it is observed that all aksharas can be formed using a maximum of 8 strokes. Hence 8 language rule models have been created. For this the combination of strokes that form a particular akshara are found out by studying the handwritten data of a large number of users. Of all the combinations only those are retained that have been used by more than 5 % writers.

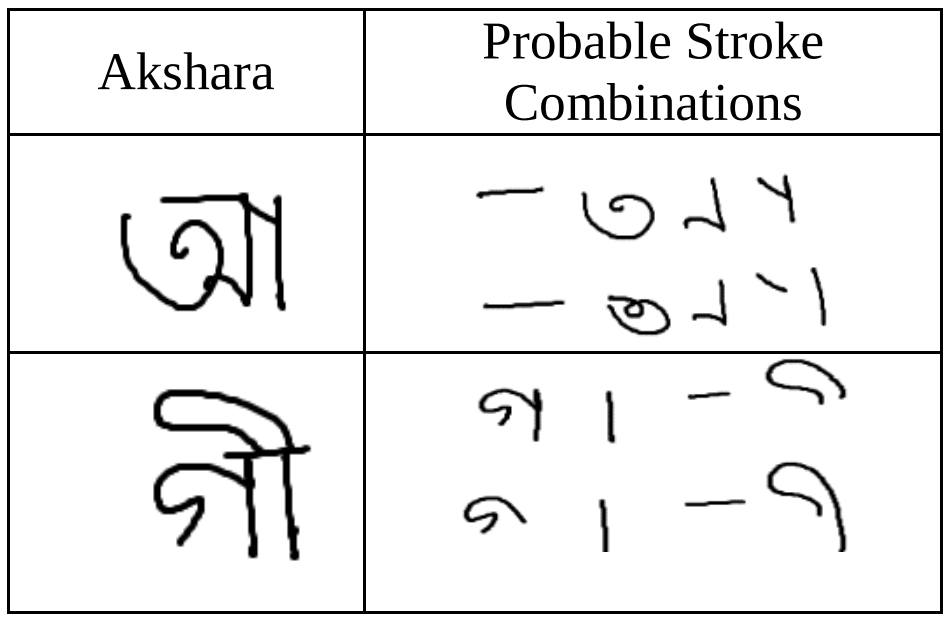

**Figure 7: Handwritten Assamese Akshara with different stroke combination**

But this level of akshara recognition gives poor recognition accuracy as the individual strokes that form an akshara regularly confuses with other strokes. So a refined language model is created which includes the combination obtained by considering the confused stroke. Confusions of individual strokes of each akshara are determined. If one stroke regularly confuses with other stroke then the new combination obtained by considering the confused stroke is added into the refined language model. Hence the final language rule model includes the combination of confused strokes as well.

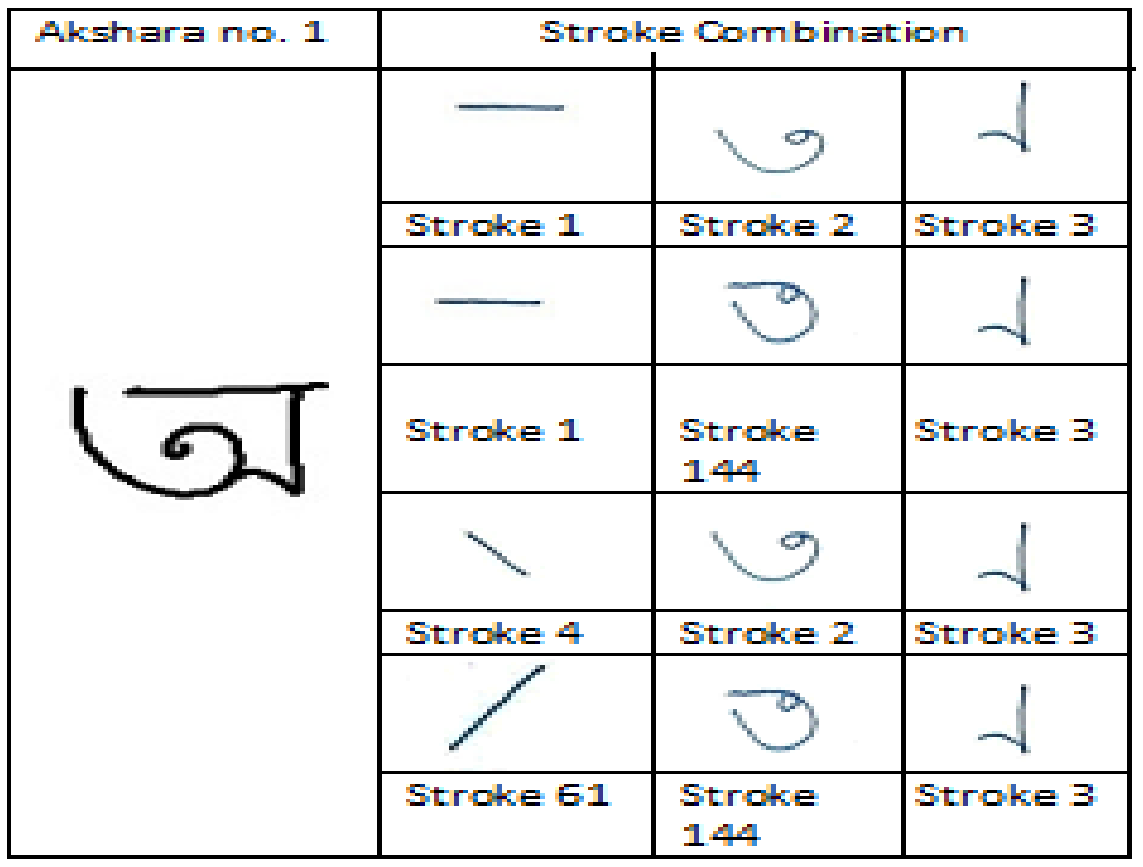

**Figure 8: Akshara 1 with primary stroke combinations and stroke confusions**

For eg, ak1 can be written as a combination of st1, st2, st3. Here st2 is confused with st144 in majority cases. And st1 is confused with st4, st61, st32, etc. hence according to the new language rule ak1 is not only a combination of st1,st2, st3 but also of (st1,st144,st3), (st4,st2,st3), (st61,st2,st3),(st4,st144,st3) and so on as shown in Figure 8

### 2.6 Graphical User Interface (GUI)

GUI of testing tool developed by Center for Development of Advanced Computing Graphics and Intelligence based Script Technology Group (CDAC), Pune, India is provided with API. We have integrated our stroke and akshara recognition system with the GUI using a dynamic linked library (DLL). Figure 9 shows the interaction of GUI with the Dynamic linked library (dll).

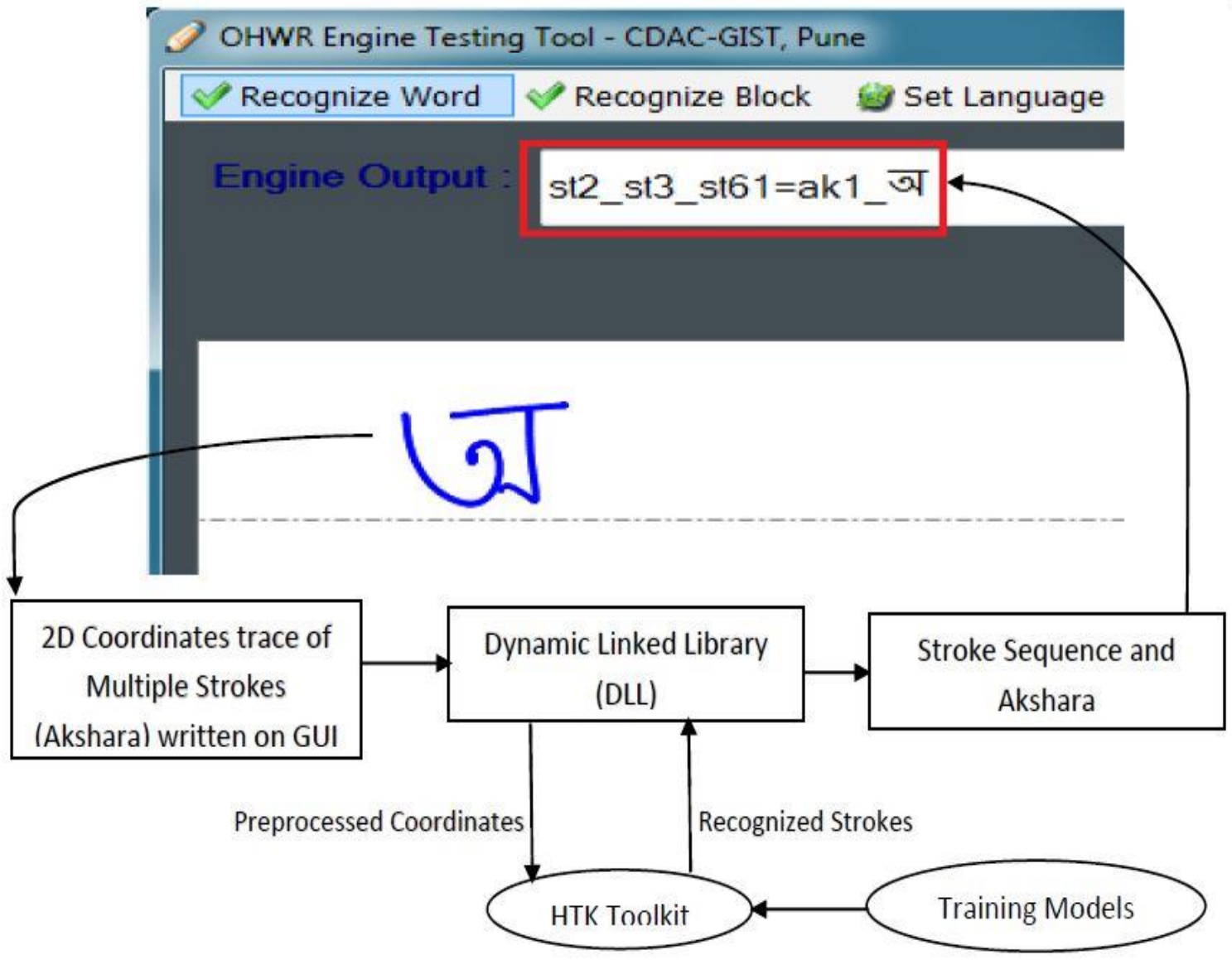

**Figure 9: Block Diagram of Akshara Recognition with GUI and DLL. Akshara 1 recognized with stroke 2, 3 and 61.**

#### 2.6.1 Akshara Recognition using Graphical User Interface

When an akshara is written on GUI the parameters of API provide the basic data like 2-dimensional coordinate traces of each stroke, the number of stroke etc. Our recognition system is integrated with the GUI using a dynamic linked library (dll). The dynamic linked library when provided with handwritten trace, performs classification tasks with HTK toolkit and delivers the recognized stroke sequence along with a Unicode character of valid akshara (if any) to the API for display. Figure 9 shows the entire process.

The essential functions performed with the dynamic linked library are as follows:

i) Preprocessing of the coordinate trace given by the API of GUI. Preprocessing involves removing duplicate points, size normalization, smoothing, interpolation of missing points & re-sampling.
ii) Features extraction.
iii) Binaries of HTK Classifier is executed with the training model file & the extracted features.
iv) The recognized strokes by HTK classifier are then read from .rec file.
v) Language rules prepared for the language are searched for a valid akshara using the set of strokes recognized by HTK classifier.
vi) Stroke sequences recognized by HTK classifier with a Unicode symbol of the akshara (if a valid akshara exists for the stroke sequence) are stored in a string data type and displayed into the GUI.

## III. EXPERIMENTAL RESULTS

For the stroke classifier, initially data is collected from 100 native writers in two sessions using a data collection tool provided by CDAC, Pune in a Tablet PC with stylus by HP Labs. Strokes from first session aksharas are used for training and strokes from second session aksharas are used for testing. Later on another two sessions of data are collected so that we have a minimum of 200 examples per akshara for training and 200 for testing. Strokes from this session of collected data are added with the previous set of data to increase the number of examples per stroke. A stroke classifier is built using HMM technique with 181 stroke classes. Hidden Markov Models are trained & tested by considering number of states as seven and Gaussian mixtures are optimized for best recognition accuracy Average recognition accuracy of the stroke classifier is 94. 91 %. The confusion matrix for the first 10 classes out of 181 classes is shown in Figure 9.

| stroke no. | st1 | st2 | st3 | st4 | st5 | st6 | st7 | st8 | st9 | st10 |
|---|---|---|---|---|---|---|---|---|---|---|
| **st1** | **69.68** | 0 | 0 | 8.30 | 0 | 1.21 | 0 | 0 | 0.75 | 0.15 |
| **st2** | 0 | **92.58** | 0 | 0 | 0 | 0 | 0 | 0 | 0 | 0 |
| **st3** | 0 | 0 | **94.67** | 0 | 0 | 0 | 0 | 0 | 0 | 0 |
| **st4** | 0.15 | 0 | 0 | **94.93** | 0 | 0.31 | 0 | 0 | 0.77 | 0 |
| **st5** | 0 | 0 | 0 | 0 | **85.71** | 0 | 0 | 0 | 0 | 0 |
| **st6** | 0.24 | 0 | 0 | 0 | 0 | **86.03** | 0 | 0 | 0.12 | 0 |
| **st7** | 0 | 0 | 0 | 0 | 0 | 0 | **95.06** | 0 | 0 | 0 |
| **st8** | 0 | 0 | 0 | 0 | 0 | 0 | 0 | **90.03** | 0 | 0 |
| **st9** | 0 | 0 | 0 | 0 | 0 | 0 | 0 | 0 | **98.31** | 0 |
| **st10** | 0 | 0 | 0 | 0 | 0 | 0 | 0 | 0 | 0 | **96.98** |

**Figure 9: Confusion matrix for the first 10 classes of stroke classifier**

For evaluation of akshara recognizer performance, users are asked to give handwritten data in the data collection tool provided by CDAC. The data saved in .xml format are annotated by which the actual class labels are attached to the akshara. A confusion matrix has been made comparing the output class label predicted by our system with actual class label using a script. The akshara level performance is 84.2% for all Assamese akshara used by our study.

## CONCLUSION

This work describes the development of online Assamese character recognition system using HMMs. A large database of handwritten Assamese numerals is collected and partitioned into two parts. One part is used for developing HMMs. The states and number of Gaussians per state are optimized by conducting large number of experiments. Finally, 7 states and 20 mixtures per state are used in the HMMs. The performance evaluation is then made for different choices of feature set, namely, only (x, y) coordinates, (x, y) coordinates and their first and second order temporal derivatives. The last case provided the best performance. Again, bottom-up approach is used in the development of akshara recognizer. First strokes corresponding to aksharas are recognized and then the aksharas. The final stroke classifier is developed for 181 distinct stroke classes and this final stroke classifier is used for Akshara recognition. The developed stroke classifier gives an average recognition accuracy of 94.14 % which is used to test combination of strokes from isolated aksharas. The akshara recognizer is evaluated by considering all strokes in an akshara. The reported recognition accuracy is 84.2 %. It is observed that, the performance is high at stroke level in comparison to akshara level performance as misclassification of the complete akshara may occur due to the single stroke mismatch. The possibility of akshara misclassification reduces with less number of strokes.

.

## AUTHORS BIOGRAPHY

**Prof. Dr. SRM Prasanna** has done his MTech from National Institute of Technology Karnataka, Surathkal in Industrial Electronics and obtained his PhD degree in CSE from Indian Institute of Technology, Madras. He is currently working as Professor in department of EEE at Indian Institute of Technology, Guwahati. His research interests include Speech Signal Processing, Speech Enhancement, Speaker Recognition, Speech Recognition, Speech Synthesis and Handwriting Recognition. E-mail: prasana@iitg.ernet.in

**Mr. Subhankar Ghosh** has completed his Master of Technology from Indian Institute of Technology, Guwahati. His research interests include Optical Character Recognition.
E-mail: ghoshsubhankar@gmail.com

**Mr. Krishna Naik** has completed his Master of Technology from Indian Institute of Technology, Guwahati. His research interest includes Handwriting Recognition.
E-mail: krishnanaik.35@gmail.com

**Ms. Rituparna Devi** has completed her Bachelor of Technology from Tezpur University, Tezpur, India and held the post of Assistant Project Engineer for one year in the project "Development of Online Handwriting Recognition System in Assamese" under the guidance of Dr. SRM Prasanna, Dept. of EEE, IIT Guwahati, India. E-mail: rituparna.sarma5@gmail.com

**Mr. Deepjoy Das** has completed his Bachelor of Technology from Tezpur University, Tezpur, India and held the post of Assistant Project Engineer for one year in the project "Development of Online Handwriting Recognition System in Assamese" under the guidance of Dr. SRM Prasanna, Dept. of EEE, IIT Guwahati, India. E-mail: deepjoy2002@gmail.com